\documentclass[11pt]{article}

\usepackage[preprint]{acl}

\usepackage{times}
\usepackage{latexsym}
\usepackage[T1]{fontenc}
\usepackage[utf8]{inputenc}
\usepackage{microtype}
\usepackage{inconsolata}
\usepackage{graphicx}
\usepackage{booktabs}
\usepackage{multirow}
\usepackage{xcolor}
\usepackage{amsmath}
\usepackage{amssymb}
\usepackage{enumitem}
\usepackage{tikz}
\usetikzlibrary{positioning, arrows.meta, fit, shapes.geometric}
\usepackage{algorithm}
\usepackage{algpseudocode}
\usepackage{tcolorbox}

\newcommand{\sogrone}{\textsc{SoG-R1}}
\newcommand{\sog}{\textsc{SoG}}
\newcommand{\sogsft}{\textsc{SoG-SFT}}
\newcommand{\sogrzero}{\textsc{SoG-R1-Zero}}

\newcommand{\tbd}{\textcolor{red}{\textbf{TBD}}}

\title{Search-on-Graph-R1: Training Large Language Models to Search Knowledge Graphs with Reinforcement Learning}

\author{
\begin{tabular}{c}
Jia Ao Sun$^{1,2}$ \quad Yu Hao$^{3,2}$ \quad Fengran Mo$^{1}$ \quad Zhan Su$^{4}$ \\
Yuchen Hui$^{1}$ \quad Bang Liu$^{1,2,5}$ \quad Jian-Yun Nie$^{1}$
\end{tabular} \\
\vspace{2pt} \\
$^{1}$Universit\'{e} de Montr\'{e}al \quad $^{2}$Mila -- Qu\'{e}bec AI Institute \\
$^{3}$McGill University \quad $^{4}$Halmstad University College \quad $^{5}$Canada CIFAR AI Chair
}

\begin{document}
\maketitle

\begin{abstract}
Knowledge graph question answering (KGQA) requires navigating from topic entities to an answer several relations away. Recent methods prompt a frontier LLM to explore the graph through a retrieval tool, but their reliance on frontier-scale inference makes them costly to deploy. We present Search-on-Graph-R1 (\sogrone{}), which internalizes this navigation into a compact 8B model through supervised fine-tuning (SFT) followed by reinforcement learning (RL). Our central idea is to scaffold a frontier teacher with each question's gold SPARQL query, so the teacher traverses a known answer-bearing path with a live \texttt{Search} tool rather than having to discover the path itself. Since every call executes against a live Freebase server, the resulting trajectories are grounded in the knowledge graph by construction. On WebQSP, CWQ, and GrailQA, \sogrone{} at 8B surpasses every frozen frontier-LLM system in our comparison and posts the strongest results on CWQ of any system we compare against. It does so using no auxiliary module at inference and no LLM judge during training. Isolating each training stage shows that SFT and RL contribute complementary gains, our approach transfers across model families, and RL learns to reach answers in fewer \texttt{Search} calls than its SFT initialization.
\end{abstract}

\section{Introduction}
\label{sec:intro}

\begin{figure*}[t]
\centering
\includegraphics[width=\textwidth]{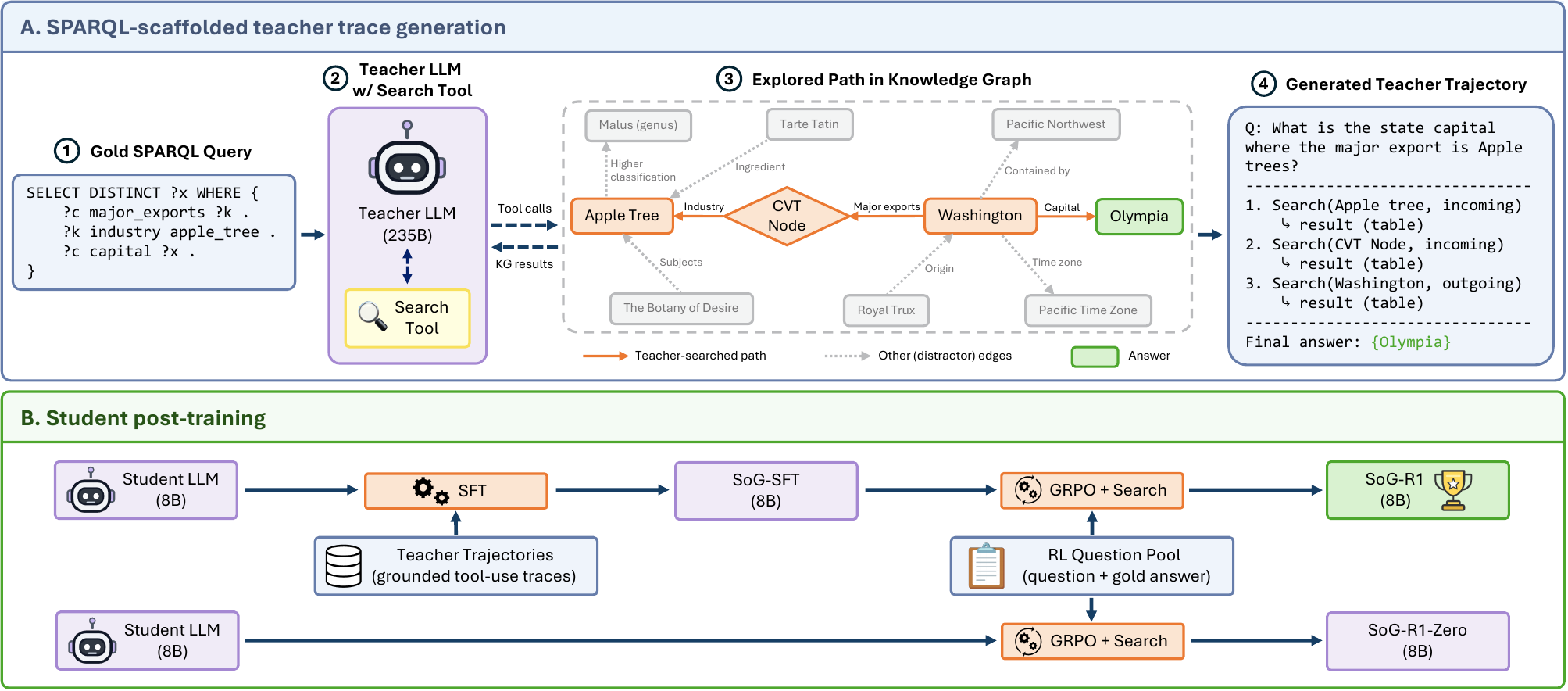}
\caption{Overview of \sogrone{}. \textbf{(A) SPARQL-scaffolded teacher trace generation.} We walk through the CWQ question \emph{``What is the state capital where the major export is Apple trees?''} \textbf{(1)} Its gold SPARQL query is shown to a 235B teacher LLM. \textbf{(2)} Using the \texttt{Search} tool, the teacher reads the query as a traversal blueprint and issues the tool calls needed to navigate the knowledge graph, executing each against a live Freebase server. \textbf{(3)} The resulting path runs \emph{Apple Tree} $\to$ CVT node $\to$ \emph{Washington} $\to$ \emph{Olympia}, the answer; distractor edges the teacher must reject are faded. \textbf{(4)} The executed calls and their results are serialized into a multi-turn teacher trajectory ending in the final answer. \textbf{(B) Student post-training.} Both branches start from the same off-the-shelf 8B instruct model. \emph{Top:} LoRA SFT on the grounded teacher trajectories of (A) produces \sogsft{}; GRPO with the same \texttt{Search} tool, initialized from \sogsft{} and trained on a pool of (\textit{question}, \textit{gold answer}) pairs, produces \sogrone{}. \emph{Bottom:} applying that same GRPO stage directly to the base 8B model, with no SFT, produces \sogrzero{}, the RL-only ablation.}

\label{fig:pipeline}
\end{figure*}

Large language models (LLMs) excel at general-purpose reasoning, but knowledge-intensive tasks remain a persistent failure mode; their parametric knowledge is incomplete or outdated, and difficult to verify~\citep{ji2023survey,de2021editing}. Pairing LLMs with knowledge graphs (KGs) offers a principled remedy~\citep{pan2024unifying}; a KG encodes explicit, structured relations between entities, allowing an LLM agent to ``hop'' across connected entities and recover multi-hop answer paths that flat semantic retrieval would miss. A growing line of work shows that frontier LLMs, by being prompted to interact with a KG through structured retrieval tools, can interleave reasoning with explicit KG queries and reach competitive accuracy on complex multi-hop KGQA benchmarks~\citep{sun2024think,chen2024plan,sun2025search}.

These prompting-based methods, however, all rely on GPT-4-class frozen frontier LLMs that are impractical for production deployment due to latency, cost, and access constraints. The next step is to post-train a smaller open-source LLM, teaching it to reason over the KG to match these frontier systems. One family of methods treats KGQA as semantic parsing, training the model to emit a complete logical form (e.g., an s-expression) that is compiled to an executable query and run against the KG~\citep{ye-etal-2022-rng,luo-etal-2024-chatkbqa,yu2023decaf}. Because the parse commits to schema-specific relations in a single pass, these methods struggle when the gold parse involves schema elements unseen at training. A second family of methods treats the KG as a black-box search environment and trains the model to use a general retrieval tool, much as Search-R1~\citep{jin2025searchr} trains LLMs to query a web search engine without memorizing its index. We adopt this framing, which yields a navigation policy over the KG rather than a fixed parse.

Two concurrent efforts adopt the same framing but keep retrieval and reasoning disjoint. EoG~\citep{yan2026exploreongraph} shares our two-stage SFT-then-RL pipeline, but extracts a per-question subgraph offline, places it in the prompt as static text, and then reasons over that fixed context without ever issuing a query; accuracy is then bounded by the recall of a preprocessing step the model cannot influence. KG-Hopper~\citep{wang2026kghopperempoweringcompactopen} does query the KG live, but its reward loop depends on external LLM judges to score reasoning traces and verify answers. Both permit the model to fall back on parametric memory when retrieval falls short, so neither guarantees that the facts it cites exist in the KG.

\sogrone{} instead fuses retrieval and reasoning into a single loop; the model itself decides what to fetch, mid-reasoning, through live \texttt{Search} calls, so that recall becomes a learnable property of navigation rather than a fixed property of preprocessing. To cold-start this behavior, we scaffold a frontier teacher with each question's gold SPARQL query, which is present in every KGQA training set but never shown to the student at inference, and have it traverse the known answer path using the same live \texttt{Search} tool the student will use. What it produces is a teacher trajectory: a sequence of executed search calls, the results they returned, and the final answer. Since every call executes against a live Freebase server, these trajectories are grounded in the KG by construction. We then fine-tune a compact 8B student on these trajectories and refine it with GRPO~\citep{shao2024deepseekmathpushinglimitsmathematical} under a simple outcome reward, with no gold-path supervision or judge in the reward loop. The resulting 8B agent surpasses every frozen frontier LLM system in our comparison on all three benchmarks, and posts the strongest CWQ accuracy of any system we compare against, including post-trained models with larger backbones, at a fraction of the inference cost.

Our contributions are:

\begin{itemize}
\item We turn the inference-time \sog{} paradigm into a learnable behavior of a single compact model, internalizing KG traversal into an 8B student's weights, with no auxiliary module and no LLM judge at any stage of training or inference.


\item We scaffold a frontier teacher with each question's gold SPARQL query, never shown to the student, so it traverses the answer path with the same live \texttt{Search} tool the student uses: every call executes against Freebase, so cold-start traces are KG-grounded by construction. We further batch the \sog{} tool over entities, removing a dropped-entity failure mode at small scale.

\item We evaluate on WebQSP, CWQ, and GrailQA, where the 8B agent outperforms every frozen frontier-LLM system and achieves the best CWQ accuracy in our comparison. Stage ablations show SFT and RL contribute complementary gains, and RL reaches answers in fewer \texttt{Search} calls than its SFT initialization on all three benchmarks.

\end{itemize}

\section{Related Work}
\label{sec:related}
We group KG-augmented LLM methods by whether the LLM that performs reasoning and answering at inference is itself post-trained.
\paragraph{Frozen-LLM methods.}
These methods keep the reasoning LLM fixed and supply KG access through prompting and inference-time orchestration: step-wise navigators query the KG hop-by-hop via beam search~\citep{sun2024think}, sub-objective planning~\citep{chen2024plan}, multi-agent question simplification~\citep{ma2025debate}, message passing~\citep{wan-etal-2025-digest}, ontology-guided paths~\citep{liu-etal-2025-ontology}, or tool calls over full reasoning history~\citep{sun2025search}, while others orchestrate frozen LLMs over training-free graph-aware constrained decoding~\citep{li-etal-2025-decoding-graphs} or off-the-shelf retrieval~\citep{li-etal-2024-framework, sui-etal-2025-fidelis, mavromatis-etal-2025-byokg}. A second line keeps the reasoning LLM frozen but trains a small dedicated helper to carry navigation or pruning: a GNN for 2-hop entity selection~\citep{wang-yu-2025-iquest}, a compact pruning or path-ranking model~\citep{dong-etal-2025-effiqa, lin-etal-2025-rje}, or an RL policy or value model~\citep{zhang-zhao-2025-collaborative, shen-etal-2025-reasoning}. A third line trains a small auxiliary LLM but still delegates the answer-producing reasoning to a frozen frontier model: GCR~\citep{pmlr-v267-luo25t} trains a small LLM to decode graph-constrained paths that a frozen GPT-4o then reasons over, PARoG~\citep{shi2026plananswerrefineongraph} trains a small Llama-3.1-8B planner whose sub-objectives a frozen GPT-4 answers and refines, and Rule-KBQA~\citep{zhang-etal-2025-rule} fine-tunes a small rule generator whose offline-mined templates guide a symbolic agent, with a frozen GPT-4 scoring each construction step. In all three lines KG competence stays outside the reasoning LLM's weights, so accuracy leans on a strong, often frontier backbone and on auxiliary modules whose interface must be hand-tuned.

\paragraph{Post-trained-LLM methods.}
Semantic-parsing methods fine-tune the LLM to emit executable logical forms grounded against the KG~\citep{luo-etal-2024-chatkbqa, feng-he-2025-rgr, tian-etal-2025-compkbqa}; these are frontier-free at execution but bounded by the training distribution, since the model commits to a full parse in one pass with no opportunity to revise against what the graph actually contains. Another family instead post-trains the LLM to traverse the KG directly, so the model that reasons at inference is the one that was trained. KG-Agent~\citep{jiang-etal-2025-kg} fine-tunes a 7B agent on synthesized tool-calling trajectories. A growing set of concurrent systems applies an SFT-then-RL pipeline to compact KG reasoners, but each relaxes a different one of three properties: retrieval over a curated KB, live querying rather than a static context, and a reward that is a deterministic function of the trajectory. Graph-R1~\citep{luo2026graphr1agenticgraphragframework} trains a multi-turn retrieval policy with GRPO, but in a GraphRAG setting, retrieving by embedding similarity over a hypergraph self-extracted from text rather than traversing a curated KB. EoG~\citep{yan2026exploreongraph} keeps retrieval and reasoning separate on a curated KB; it reasons over a statically extracted subgraph placed into the prompt and never issues a query, so the triples it cites are not guaranteed to exist in the KG. KG-Hopper~\citep{wang2026kghopperempoweringcompactopen} equips its agent with a live search tool, but its reward loop depends on external LLM judges---a 70B scorer that grades each reasoning trace and a 3B model that verifies answers---making its training signal stochastic and model-dependent. Both EoG and KG-Hopper permit the model to fall back on parametric memory when retrieval falls short, so neither guarantees KG-grounded answers. Our \sogrone{} instead fuses retrieval and reasoning in one compact model: retrieval is the sole knowledge channel with no parametric fallback permitted, and the reward is a deterministic function of the trajectory, with no LLM judge in the reward loop and no cooperating model at inference.

\section{Knowledge Graphs and KGQA}
\label{sec:task}
A knowledge graph is a set of fact triples $\mathcal{G} = \{(e, r, e') \mid e, e' \in \mathcal{E},\ r \in \mathcal{R}\}$, where $\mathcal{E}$ and $\mathcal{R}$ are the entity and relation sets and each triple $(e, r, e')$ states that relation $r$ holds between entities $e$ and $e'$. Knowledge Graph Question Answering (KGQA) aims to answer a natural-language question $q$ using $\mathcal{G}$; the goal is to identify the entity set $\mathcal{A}_q \subseteq \mathcal{E}$ that answers $q$. Following prior work~\citep{sun2024think, chen2024plan, sun2025search}, we assume the topic entities $\mathcal{E}_q \subseteq \mathcal{E}$ mentioned in $q$ have already been identified and linked to nodes in $\mathcal{G}$, and serve as the starting points for reasoning. Answering $q$ is rarely a one-hop lookup; the answer typically lies several relations away from the topic entities, so $\mathcal{A}_q$ is only reachable through a multi-hop path in $\mathcal{G}$, which the agent must discover by traversing the graph from $\mathcal{E}_q$.

\section{Methodology}
\label{sec:method}
\sogrone{} is built in three sequential stages applied to a compact instruction-tuned student model. \textbf{Stage 1: SPARQL-Scaffolded Cold-Start Trace Generation} (\S\ref{sec:teacher-traces}). For each training question, a frontier-scale teacher LLM walks the gold answer path using the same \texttt{Search} tool the student will use at inference, emitting a multi-turn trajectory of tool calls and observations terminating in a final answer, with the gold SPARQL query as the scaffold. \textbf{Stage 2: Supervised Fine-Tuning} (\S\ref{sec:lora}). The student is LoRA-fine-tuned on the collected trajectories, producing \sogsft{}. \textbf{Stage 3: Reinforcement Learning} (\S\ref{sec:rl-reward}). \sogsft{} is further trained with GRPO under a reward combining exact-match correctness with a turn-count efficiency factor, producing \sogrone{}.
At inference the student receives only the question $q$ and an entity dictionary $\mathcal{E}_q$ (no SPARQL) and produces an answer through the tool-augmented loop of Algorithm~\ref{alg:inference}.

\begin{algorithm}[t]
\small
\caption{\sogrone{} inference.}
\label{alg:inference}
\begin{algorithmic}[1]
\Require question $q$, topic entities $\mathcal{E}_q$, policy $\pi$, turn budget $T$
\Ensure predicted answer set $\hat{\mathcal{A}}_q$
\State $h \gets [q,\, \mathcal{E}_q]$ \Comment{running context}
\For{$t = 1, \ldots, T$}
  \State $a_t \sim \pi(\cdot \mid h)$ \Comment{sample next action}
  \If{$a_t$ is a final answer}
    \State \Return answer set $\hat{\mathcal{A}}_q$ parsed from $a_t$
  \Else \Comment{$a_t$ is a \texttt{Search} call}
    \State $o_t \gets \textsc{Search}(a_t)$
    \State $h \gets h \,\Vert\, [a_t,\, o_t]$ \Comment{append call and observation}
  \EndIf
\EndFor
\State \Return $\varnothing$ \Comment{turn budget exhausted}
\end{algorithmic}
\end{algorithm}

\subsection{Tool: Batched 1-Hop Neighbour Retrieval}
\label{sec:tool}

The agent has access to a single tool, \texttt{Search}:

\begin{tcolorbox}[colback=gray!5,colframe=gray!50,boxrule=0.5pt,arc=2pt,left=4pt,right=4pt,top=2pt,bottom=2pt]
\small
\begin{verbatim}
Search(
  entities: List[str],
  direction: "incoming" | "outgoing",
  properties: List[str] = None
) -> str
\end{verbatim}
\end{tcolorbox}

\noindent Given a list of entity IDs and a direction, \texttt{Search} returns the adjacent triples as a markdown table of properties, neighbouring entities, and their labels, each row tagged with its source entity. The optional \texttt{properties} argument restricts the result to a given set of predicates. The key change from the original \sog{}~\citep{sun2025search} tool is that it accepts a list of entities in a single call, whereas the original \texttt{Search} took only one entity at a time. When the agent needs to expand several entities from the same hop---for example the compound-value-type (CVT) nodes Freebase returns for an actor's individual film performances---the single-entity tool forces a separate call per entity. We find the model often issues calls for only a subset before moving on, dropping the rest and losing the answers behind them. Bundling the entities into one list and passing that list to \texttt{Search} makes this a single decision; the agent collects every entity it intends to expand into one call, which is far harder to truncate than a sequence of independent calls. As a secondary benefit, it also lowers the total number of tool calls per question, reducing inference cost and the chance of mid-trajectory derailment.

\subsection{Stage 1: SPARQL-Scaffolded Cold-Start Trace Generation}
\label{sec:teacher-traces}

Cold-start data for tool-using KG agents requires multi-turn trajectories that interleave reasoning with valid tool calls and reach the gold answer. Generic distillation pipelines (e.g., Toolformer~\citep{schick2023toolformer}, ToolLLM~\citep{qin2024toolllm}) prompt a teacher with the question alone and filter post-hoc by output correctness. We instead exploit a structural property of KGQA training data; every training question comes with a gold SPARQL query, which is not merely an executable that returns the answer but an explicit encoding of how to reach it in the graph. Each triple in its WHERE clause is a single hop from the topic entity in $\mathcal{E}_q$, through intermediate CVT nodes or relation endpoints, to the SELECT target that resolves to the answer---exactly the traversal a \texttt{Search}-equipped agent must produce. We therefore use the gold query to scaffold a frontier teacher; rather than asking it to rediscover a path the dataset already encodes, we give it $\sigma_q$ as a hint and have it walk that path using the same \texttt{Search} tool the student uses at inference. Every call executes against a live Freebase Virtuoso server so that each KG fact in a trajectory originates from a real query rather than model-generated prose. Figure~\ref{fig:pipeline}(A) shows an example. Topic entities are given as $\mathcal{E}_q$, as is standard for these benchmarks and consistent with our baselines. We retain a trajectory only if its final answer exactly matches the gold answer and all of its \texttt{Search} calls executed without error; this retains 99.6\% of trajectories on WebQSP, 96.0\% on CWQ, and 87.2\% on GrailQA. The full prompt is in Appendix~\ref{sec:appendix-tool}.

\subsection{Stage 2: Supervised Fine-Tuning}
\label{sec:lora}

Each retained teacher trajectory (\S\ref{sec:teacher-traces}) is serialized into a multi-turn conversation whose assistant turns are \texttt{Search} tool calls and whose intervening turns carry the returned observations, terminating in the final-answer turn. We LoRA-fine-tune~\citep{hu2022lora} the student on these conversations, computing the next-token loss only on the assistant turns, so it learns to produce tool calls and answers without being trained to predict tool outputs. We train a separate specialist per dataset, matching the few-shot trace distribution to the target benchmark and isolating the cold-start contribution per benchmark. Adapter, optimization, and hardware settings are in \S\ref{sec:appendix-training}.

\subsection{Stage 3: Reinforcement Learning}
\label{sec:rl-reward}

Starting from \sogsft{}, we apply GRPO~\citep{shao2024deepseekmathpushinglimitsmathematical} with the student interacting live with the KG. The reward and its design are below; rollout, optimization, and capacity settings are in \S\ref{sec:appendix-training}.

\paragraph{Reward.}
For a trajectory $\tau$ with solution string $s$ and $n_t$ tool calls, the reward is the product of three factors:
\begin{equation*}
  r(\tau) = g(\tau) \cdot r_{\mathrm{em}}(\tau) \cdot f(n_t),
\end{equation*}
a binary format gate $g(\tau)\in\{0,1\}$, a binary exact-match term $r_{\mathrm{em}}(\tau)\in\{0,1\}$, and a turn-count efficiency factor $f(n_t)$. The format gate $g(\tau)=1$ iff (i) $n_t \geq 1$, so the model must call the tool at least once, ruling out direct-answer rollouts that bypass KG navigation, and (ii) $s$ terminates cleanly with a single "\texttt{Final answer:}", rejecting rollouts that wrap the answer in commentary or continue generating past it, both common in early RL. The exact-match term $r_{\mathrm{em}}(\tau)$ returns $1$ if the predicted answer set matches the gold answer and $0$ otherwise; there is no partial credit giving GRPO a sparse but unambiguous correctness signal. The turn-count factor penalizes verbosity through a floored linear penalty on excess turns,
\begin{equation*}
  f(n_t) = \max\!\bigl(f_{\min},\; 1 - \lambda\cdot\max(0,\, n_t - T_d)\bigr),
\end{equation*}
governed by three constants: a free-zone threshold $T_d$, below which no penalty applies; a per-turn penalty rate $\lambda$; and a floor $f_{\min}$ that caps the maximum penalty. Crucially, the factor is multiplicative, so efficiency pressure operates only among correct rollouts; since $r_{\mathrm{em}}=0$ on any incorrect trajectory, $f(n_t)$ acts as a tie-breaker that prefers shorter paths without ever losing correctness. Neither $\lambda$ nor $f_{\min}$ can change the sign of the advantage between a correct and an incorrect rollout, only the margin between two correct ones. We therefore fix both at conservative values and do not tune them.

\section{Experiments}
\label{sec:experiments}

We organize our evaluation around five research questions. \textbf{RQ1:} How does \sogrone{}, at 8B, compare against frozen frontier LLM and post-trained KGQA systems? \textbf{RQ2:} What does each post-training stage (SFT cold-start, RL) contribute, and is cold-start necessary for RL to work at all? \textbf{RQ3:} Does the RL stage learn fewer search calls than its SFT initialization? \textbf{RQ4:} How does test accuracy scale with the size of the grounded training pool? \textbf{RQ5:} Does our approach transfer across different base models and parameter sizes?

\subsection{Setup}

\paragraph{Datasets.}
We evaluate on three standard Freebase~\citep{bollacker2008freebase} KGQA benchmarks: \textbf{WebQSP}~\citep{yih-etal-2016-value}, \textbf{CWQ}~\citep{talmor-berant-2018-web}, and \textbf{GrailQA}~\citep{gu2021beyond}.

\paragraph{Metrics.}
Following prior work~\citep{li-etal-2023-shot, sun2024think, chen2024plan}, we report exact match accuracy (Hits@1) over the answer set.

\paragraph{Baselines.}
We compare against three families, reflecting the taxonomy of \S\ref{sec:related}. \emph{LLM-only} baselines (IO prompting, Chain-of-Thought~\citep{wei2022chain}, Self-Consistency~\citep{wang2022self}) run the same 8B backbone with no KG access, establishing the no-KG floor for this backbone. The following \emph{frozen-LLM} KG methods pair a frozen frontier LLM with inference-time orchestration: ToG~\citep{sun2024think}, PoG~\citep{chen2024plan}, LMP~\citep{wan-etal-2025-digest}, iQUEST~\citep{wang-yu-2025-iquest}, ORT~\citep{liu-etal-2025-ontology}, DoG~\citep{ma2025debate}, PARoG~\citep{shi2026plananswerrefineongraph}, and the prompted \sog{}~\citep{sun2025search}. The following \emph{post-trained} methods fine-tune the reasoning LLM on KGQA data: ChatKBQA~\citep{luo-etal-2024-chatkbqa}, Reasoning-on-Graph~\citep{luo2024reasoning}, KG-Agent~\citep{jiang-etal-2025-kg}, RGR-KBQA~\citep{feng-he-2025-rgr}, CompKBQA~\citep{tian-etal-2025-compkbqa}, D-RAG~\citep{gao-etal-2025-rag}, KG-Hopper~\citep{wang2026kghopperempoweringcompactopen}, and EoG~\citep{yan2026exploreongraph}. Of these, EoG and KG-Hopper are the concurrent systems most directly comparable to ours; both post-train Llama-3.1-8B, our exact backbone.

\paragraph{Training and validation data.}
Within each dataset, SFT and RL operate on the same retained teacher-trace pool (\S\ref{sec:teacher-traces}), so the gap between \sogsft{} and \sogrone{} reflects the training procedure rather than a difference in the questions each stage draws on, and each stage holds out a fixed 300-question validation set. We train on 8{,}700 questions for CWQ and 2{,}700 for GrailQA; for WebQSP, whose retained pool is an order of magnitude smaller, we use all 2{,}707 available after the validation split. GrailQA also draws on a smaller pool because its class-level questions, whose gold SPARQL binds no topic entity, leave entity-seeded \texttt{Search} with no starting point and are filtered out. Table~\ref{tab:data-counts} gives the per-stage counts, and \S\ref{sec:ablation-data-scale} quantifies the effect of pool size.

\begin{table}[t]
  \centering
  \small
  \begin{tabular}{lcccc}
    \toprule
    & \multicolumn{2}{c}{\textit{SFT}} & \multicolumn{2}{c}{\textit{RL}} \\
    \cmidrule(lr){2-3}\cmidrule(lr){4-5}
    Dataset & train & val & train & val \\
    \midrule
    WebQSP  & 2{,}707 & 300 & 2{,}707 & 300 \\
    CWQ     & 8{,}700 & 300 & 8{,}700 & 300 \\
    GrailQA & 2{,}700 & 300 & 2{,}700 & 300 \\
    \bottomrule
  \end{tabular}
  \caption{Training and validation question counts per stage. SFT and RL draw from the same pool per dataset; WebQSP uses its entire retained teacher-trace pool.}
  \label{tab:data-counts}
\end{table}

\paragraph{Implementation.} The student is Llama-3.1-8B-Instruct~\citep{grattafiori2024llama3herdmodels} and the cold-start teacher is Qwen3-235B-A22B-Thinking-2507 (FP8)~\citep{yang2025qwen3}. SFT uses LoRA via LLaMA-Factory~\citep{zheng-etal-2024-llamafactory}, and RL uses GRPO in verl~\citep{sheng2024hybridflow} with vLLM~\citep{kwon2023efficient} rollouts, all on a single $8{\times}$H200 141GB node. Both teacher trace generation and RL rollouts dispatch \texttt{Search} calls to a self-hosted Freebase Virtuoso instance. For the turn-count factor we fix $\lambda = 0.05$ and $f_{\min} = 0.3$ across all datasets; the free-zone threshold $T_d$ is set per dataset to the 75th-percentile turn count of that dataset's teacher trajectories. Full hyperparameters and per-stage walltimes are in Appendix~\ref{sec:appendix-training}.

\begin{table*}[t]
  \centering
  \small
  \begin{tabular}{llcccc}
    \toprule
    \textbf{Method} & \textbf{Backbone} & \textbf{Size} & \textbf{WebQSP} & \textbf{CWQ} & \textbf{GrailQA} \\
    \midrule
    \multicolumn{6}{c}{\textit{LLM-only}} \\
    \midrule
    IO Prompt                       & Llama-3.1 & 8B     & 62.2 & 38.3 & 24.6 \\
    Chain-of-Thought                & Llama-3.1 & 8B     & 60.2 & 41.4 & 28.8 \\
    Self-Consistency                & Llama-3.1 & 8B     & 61.2 & 42.0 & 26.6 \\
    \midrule
    \multicolumn{6}{c}{\textit{Frozen-LLM Methods}} \\
    \midrule
    Think-on-Graph~\citep{sun2024think}    & GPT-4   & --  & 82.6 & 69.5 & 81.4 \\
    Plan-on-Graph~\citep{chen2024plan}     & GPT-4   & --  & 87.3 & 75.0 & 84.7 \\
    LMP~\citep{wan-etal-2025-digest}       & GPT-4   & --  & 90.0 & 82.2 & 89.3 \\
    iQUEST~\citep{wang-yu-2025-iquest}     & GPT-4o  & --  & 88.9 & 73.9 & 73.5 \\
    ORT~\citep{liu-etal-2025-ontology}     & GPT-4o  & --  & 87.7 & 65.4 & --   \\
    Debate-on-Graph~\citep{ma2025debate}   & GPT-4   & -- & 91.0  & 56.0 & 80.0 \\
    PARoG~\citep{shi2026plananswerrefineongraph} & GPT-4 & -- & 91.2 & 79.3 & 87.1 \\
    Search-on-Graph~\citep{sun2025search}           & GPT-4o  & --  & 91.3 & 75.1 & 86.9 \\
    \midrule
    \multicolumn{6}{c}{\textit{Post-trained LLM Methods}} \\
    \midrule
    ChatKBQA~\citep{luo-etal-2024-chatkbqa}        & Llama-2       & 13B & 86.4 & \underline{86.0} & --   \\
    Reasoning-on-Graph~\citep{luo2024reasoning}    & Llama-2-Chat  & 7B  & 85.7 & 62.6 & --   \\
    KG-Agent~\citep{jiang-etal-2025-kg}            & Llama-2       & 7B  & 83.3 & 72.2 & 86.1 \\
    RGR-KBQA~\citep{feng-he-2025-rgr}              & Llama-3       & 8B  & 84.5 & 82.0 & --   \\
    CompKBQA~\citep{tian-etal-2025-compkbqa}       & Llama-2       & 13B & 84.2 & 83.6 & --   \\
    D-RAG~\citep{gao-etal-2025-rag}                & Llama-3-Inst. & 8B  & 89.1 & 70.3 & --   \\
    KG-Hopper~\citep{wang2026kghopperempoweringcompactopen} & Llama-3.1 & 8B & 76.9 & 58.2 & 55.4 \\
    Explore-on-Graph~\citep{yan2026exploreongraph} & Llama-3.1     & 8B  & \textbf{92.8} & 82.6 & \textbf{92.1} \\
    \midrule
    \textbf{\sogrone{} (Ours)}                     & Llama-3.1 & 8B & \underline{91.7} & \textbf{88.7} & \underline{90.0} \\
    \bottomrule
  \end{tabular}
  \caption{Exact match accuracy (\%) on WebQSP, CWQ, and GrailQA, grouped by the taxonomy of \S\ref{sec:related}: LLM-only (no KG), frozen-LLM methods, and post-trained LLM methods. Baseline numbers are taken from the cited papers' best-reported configurations. Bold marks the best result in each column, underline the second-best. ``--'' indicates a dataset the method does not report on.}

  \label{tab:main}
\end{table*}

\subsection{Main Results}
\label{sec:main-results}

Table~\ref{tab:main} compares \sogrone{} against frozen frontier-LLM and post-trained systems. Three findings stand out.

First, \sogrone{} exceeds every frozen frontier system on all three benchmarks---well above its KG-free floor of $62.2$/$42.0$/$28.8$ (the same Llama-3.1-8B backbone with no KG access) and above the strongest frozen entries, including \sog{} ($91.3$/$75.1$/$86.9$) and LMP (GPT-4) on CWQ and GrailQA. The margin is decisive on CWQ ($+6.5$ over LMP) but narrower on WebQSP ($+0.4$) and GrailQA ($+0.7$).

Second, among post-trained methods \sogrone{} posts the best CWQ accuracy of any system in the table ($88.7$, over ChatKBQA's $86.0$) and is second to EoG on WebQSP and GrailQA. This comparison is backbone-matched---both post-train Llama-3.1-8B---but the two systems are given different inputs at inference. EoG reasons over a per-question subgraph produced by a separate, offline retrieval step and placed in its prompt as static text; its policy issues no queries, so the region of the KG available to the model is fixed before it runs and recall is a property of that external step rather than of the learned policy. \sogrone{} is only given the topic entities and must locate the answer region itself through live \texttt{Search} calls over the full KG. The results therefore cannot be compared directly; \sogrone{}'s $6.1$-point CWQ lead is obtained while also performing retrieval, and the $1.1$ and $2.1$ point gaps on WebQSP and GrailQA are small relative to that lead.

Third, against KG-Hopper, which shares our backbone and SFT-then-RL structure, \sogrone{} gains $14.8$/$30.5$/$34.6$ points. Overall, \sogrone{} delivers the strongest CWQ accuracy of any system in the table with a single 8B policy, with no judge or path supervision in training.

\subsection{Training Stage Ablation}
\label{sec:ablation-stages}

To isolate each post-training stage, we compare four configurations of the same Llama-3.1-8B-Instruct backbone, evaluated on all three test sets under an identical \texttt{Search} tool, system prompt, and decoding setup:
\begin{itemize}
\item \textbf{Base} (zero-shot inference): the off-the-shelf instruct model, no fine-tuning, prompted with the SoG system prompt and the \texttt{Search} tool.
\item \textbf{\sogsft{}} (SFT only): LoRA SFT on SPARQL-scaffolded 235B-teacher traces.
\item \textbf{\sogrzero{}} (RL only): GRPO applied directly to the base instruct model with no cold-start data, with the same reward, same hyperparameters as the full pipeline.
\item \textbf{\sogrone{}} (full pipeline, SFT then RL): the proposed system. SFT on 235B-teacher traces, then GRPO with the \texttt{Search} tool.
\end{itemize}

\begin{table}[t]
  \centering
  \small
  \resizebox{\columnwidth}{!}{%
  \begin{tabular}{lccc}
    \toprule
    Configuration            & WebQSP & CWQ    & GrailQA \\
    \midrule
    Base (zero-shot)         &  50.9  &  35.3  &  55.5 \\
    \sogrzero{} (RL only)    &  80.9  &  63.0  &  60.7 \\
    \sogsft{} (SFT only)     &  89.6  &  86.1  &  88.8 \\
    \sogrone{} (SFT + RL)    & \textbf{91.7}  &  \textbf{88.7}  &  \textbf{90.0} \\
    \bottomrule
  \end{tabular}%
  }
  \caption{Effect of each post-training stage on the same Llama-3.1-8B-Instruct backbone, reported as mean\,$\pm$\,standard deviation over three seeds. Base is a single deterministic zero-shot run. \sogrzero{} tests whether SFT warm-up is necessary; \sogsft{} isolates the cold-start contribution. The gap from \sogsft{} to \sogrone{} is the uplift RL adds on top of SFT; the gap from \sogrzero{} to \sogrone{} is what SFT warm-up adds on top of RL alone.}
  \label{tab:ablation-stages}
\end{table}

Table~\ref{tab:ablation-stages} shows the contribution of each stage. We make three observations. (i) SFT contributes the largest single jump: cold-start alone lifts the model to $90.9$/$86.3$/$88.8$, a gain of $+41.7$/$+54.1$/$+34.7$ over Base, confirming that the SPARQL-scaffolded traces transfer a strong navigation prior. (ii) RL further improves SFT policy: the full pipeline reaches $91.7$/$88.7$/$90.0$, improving over SFT-only by $+0.8$/$+2.4$/$+1.2$. The uplift is largest on CWQ, the deepest multi-hop benchmark, consistent with RL primarily teaching the recovery and re-search behaviors that matter most when a single canonical path is insufficient. (iii) RL-only is worse than SFT-only: GRPO applied directly to the base model (\sogrzero{}) reaches only $81.9$/$63.0$/$60.0$, below SFT-only on every benchmark, and the ordering of $\textsc{SoG-R1-Zero} < \textsc{SoG-SFT} < \textsc{SoG-R1}$ in regards to performance.

\subsection{Path Compression: RL Discovers Shorter Paths than SFT}
\label{sec:path-compression}

A central claim of \sogrone{} is that the RL stage teaches the agent to navigate the KG more concisely than its SFT initialization. We measure trajectory compactness as the average number of \texttt{Search} tool calls per question, computed on correctly-answered test rollouts (Table~\ref{tab:compression}).

\begin{table}[t]
  \centering
  \small
  \resizebox{\columnwidth}{!}{%
  \begin{tabular}{lccc}
    \toprule
    Configuration              & WebQSP & CWQ & GrailQA \\
    \midrule
    \sogsft{} (SFT only)       & 2.68 & 4.98 & 2.08   \\
    \sogrone{} (SFT + RL)      & \textbf{2.28} & \textbf{4.80} & \textbf{1.88}   \\
    \bottomrule
  \end{tabular}%
  }
  \caption{Average \texttt{Search} calls per correctly-answered question. RL shortens trajectories relative to its SFT initialization on all three benchmarks while accuracy rises (Table~\ref{tab:ablation-stages}), so the compression is not bought with correctness.}
  \label{tab:compression}
\end{table}

The RL stage shortens paths on every benchmark: WebQSP $2.68 \to 2.28$ ($-15\%$), CWQ $4.98 \to 4.80$ ($-4\%$), and GrailQA $2.08 \to 1.88$ ($-10\%$). Accuracy rises in all three cases, by $+0.8$/$+2.4$/$+1.2$ EM (Table~\ref{tab:ablation-stages}), so the agent is not reaching answers in fewer tool calls by abandoning the questions that need more.

This shortening is the opposite of the response-length growth reported in Search-R1's web-search setting~\citep{jin2025searchr}; where their agent learns to issue more search calls over training, ours learns to issue fewer. Each KG \texttt{Search} is costly in latency and context, and the turn-count factor rewards reaching the answer in fewer hops among correct trajectories.

\subsection{Data-Scale Ablation}
\label{sec:ablation-data-scale}
WebQSP's main configuration already uses every retained trajectory, but CWQ and GrailQA retain far more than the pipeline consumes, leaving headroom to vary the pool size. We re-run the full SFT-then-RL pipeline on CWQ at increasing pool sizes (only the pool grows, with both stages drawing from it and all else held fixed) and report each stage separately. GrailQA is omitted; its EM was unchanged from $2{,}700$ to $8{,}700$ questions, so it is already saturated at the main configuration.
\begin{figure}[t]
  \centering
  \includegraphics[width=\columnwidth]{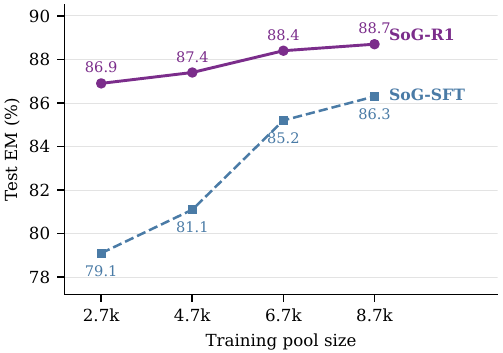}
  \caption{Data-scale ablation on CWQ: test-set EM as the post-training pool grows, for the cold-start stage alone (\sogsft{}) and for the full SFT-then-RL pipeline (\sogrone{}). Both stages draw from the same pool at every size, and the backbone, tool, prompts, validation-set size, and all hyperparameters are held fixed. The $8{,}700$ point is the main \sogrone{} configuration of Table~\ref{tab:main}; the $2{,}700$ point matches the pool size of the smallest dataset (WebQSP).}
  \label{fig:data-scaling}
\end{figure}

Figure~\ref{fig:data-scaling} separates the two stages. \sogsft{}'s EM climbs steeply, from $79.1$ to $86.3$ ($+7.2$), and is still rising at $8{,}700$ questions. \sogrone{} rises from $86.9$ to $88.7$ ($+1.8$), plateauing by $8{,}700$ where the final $2{,}000$ questions add only $+0.3$.

RL compensates for scarce cold-start data. The uplift RL adds over its initialization shrinks as the pool grows, from $+7.8$ EM at $2{,}700$ questions to $+2.4$ at $8{,}700$. With a small pool, RL recovers most of the accuracy SFT alone cannot reach; with a large pool, SFT is already strong and RL adds little. This matters most where annotation is scarce. WebQSP's entire pool is only $2{,}707$ questions, exactly where RL is most valuable.

The pipeline also plateaus early. The last 2{,}000 CWQ questions add only +0.3 EM, so we never use the full 24{,}680-trajectory pool. Combined with the high trace retention of \S\ref{sec:teacher-traces} (96--99\%), this means only a few thousand annotated questions, distilled once, are needed to come close to the best accuracy this approach can achieve.

\subsection{Backbone Generality}
\label{sec:ablation-backbone}

Our main results use a single backbone, leaving open whether \sogrone{}'s gains come from the training approach or from one model family's tool-use ability. We re-run the full pipeline on Qwen3-4B-Instruct-2507~\citep{yang2025qwen3}, a different family at roughly half the scale.

\begin{table}[t]
  \centering
  \small
  \resizebox{\columnwidth}{!}{%
\begin{tabular}{llccc}
    \toprule
    Backbone & Stage & WebQSP & CWQ & GrailQA \\
    \midrule
    \multirow{3}{*}{Llama-3.1-8B-Inst}  & Base       & 49.2 & 32.2 & 54.1 \\
                                        & \sogsft{}  & 90.9 & 86.3 & 88.8 \\
                                        & \sogrone{} & \textbf{91.7} & 88.7 & \textbf{90.0} \\
    \midrule
    \multirow{3}{*}{Qwen3-4B-Inst}      & Base       & 55.5 & 32.5 & 60.9 \\
                                        & \sogsft{}  & 89.4 & 82.9 & 86.8 \\
                                        & \sogrone{} & 91.5 & \textbf{89.4} & 88.1 \\
    \bottomrule
\end{tabular}
  }
\caption{Backbone generality. The full pipeline is re-run on Qwen3-4B-Instruct-2507 with identical data, tool, prompt, and hyperparameters. Base is the untrained instruction model with the \texttt{Search} tool; \sogsft{} is the SFT-only stage and \sogrone{} the full pipeline. Llama-3.1-8B rows repeat Table~\ref{tab:ablation-stages}.}

  \label{tab:ablation-backbone}
\end{table}

Table~\ref{tab:ablation-backbone} reports the comparison and we make two observations here.

(i) The approach transfers across model families. On both backbones the ordering of $\textsc{Base} < \textsc{SoG-SFT} < \textsc{SoG-R1}$ holds and the shape is the same. SPARQL-scaffolded SFT sets a strong prior and RL adds a smaller refinement. Neither stage depends on model family.

(ii) RL contributes more when the SFT initialization is weaker. On Qwen3-4B, RL adds $+2.1$/$+6.5$/$+1.3$ EM over SFT, against $+0.8$/$+2.4$/$+1.2$ on Llama-3.1-8B. The stronger backbone reaches a higher SFT plateau, leaving RL less to recover. This is the same substitution seen in \S\ref{sec:ablation-data-scale}, where the RL uplift shrinks as the cold-start pool grows.

We nonetheless headline Llama-3.1-8B for a reason unrelated to accuracy; the closest concurrent systems, EoG~\citep{yan2026exploreongraph} and KG-Hopper~\citep{wang2026kghopperempoweringcompactopen}, both post-train it, so matching the backbone makes those comparisons controlled for base model.

\section{Conclusion}

We present \sogrone{}, a two-stage post-training pipeline that internalizes live KG navigation into a compact 8B agent. We scaffold a frontier teacher with each question's gold SPARQL query so it traverses the answer path with a live \texttt{Search} tool, producing cold-start trajectories that are grounded in the KG by construction. The student is then fine-tuned on these trajectories and refined with GRPO under a deterministic outcome reward. On WebQSP, CWQ, and GrailQA, the resulting agent surpasses every frozen frontier-LLM system in our comparison and posts the strongest CWQ accuracy of any system we compare against. Ablations show SFT and RL contribute complementary gains and the approach transfers across model families. RL also makes navigation more efficient, reaching answers in fewer \texttt{Search} calls than its SFT initialization.

\section*{Limitations}
\label{sec:limitations}

We evaluate \sogrone{} only on Freebase, the KG underlying WebQSP, CWQ, and GrailQA, and our pipeline assumes a KG that exposes 1-hop neighbour retrieval and a per-question gold SPARQL query at training time. Whether SPARQL-scaffolded grounding transfers to graphs without such annotations (e.g., Wikidata) is untested. We further follow standard practice for these benchmarks in providing gold topic entities (oracle entity linking): the student receives the linked entity IDs $\mathcal{E}_q$ rather than performing entity linking itself, and performance under self-linking is presumed to be lower.


\bibliography{custom}

\appendix

\section{Tool Definition}
\label{sec:appendix-tool}

\paragraph{Tool schema.}
The student and teacher both interact with the KG through a single tool, \texttt{Search}, defined in OpenAI function-calling JSON schema:
\begin{small}
\begin{verbatim}
{
  "name": "Search",
  "description": "Execute a SPARQL query on
   Freebase that retrieves adjacent
   properties, property labels, values, and
   value labels in the specified direction
   for one or more entities. Supports batch
   querying: pass a list of entity IDs to
   query multiple entities at once.",
  "parameters": {
    "type": "object",
    "properties": {
      "entities": {
        "type": "array",
        "description": "List of entity IDs
         to query (e.g. ['m.04yd0fh'] for a
         single entity; ['m.02tbdk9',
         'm.0jy1jy'] for batch). Use batch
         querying when querying multiple
         entities with the same property
         filter."
      },
      "direction": {
        "type": "string",
        "enum": ["incoming", "outgoing"],
        "description": "'outgoing' for
         relations where the entity is the
         subject; 'incoming' for relations
         where the entity is the object."
      },
      "properties": {
        "type": "array",
        "description": "Optional list of
         specific properties to filter by
         (e.g. ['film.actor.film']). Use to
         reduce results when too many are
         returned."
      }
    },
    "required": ["entities", "direction"]
  }
}
\end{verbatim}
\end{small}

The executor parses each call, dispatches a SPARQL query to a Freebase Virtuoso server, and returns a Markdown table of \texttt{(property, propertyLabel, value, valueLabel)} rows; for batched calls over multiple entities, each row additionally carries \texttt{entity} and \texttt{entityLabel} columns identifying the input entity it came from. When the result exceeds 50 rows, the executor instead returns a paged property summary that prompts the agent to retry with a narrowed \texttt{properties}.

\paragraph{System prompts.}
The teacher and student use different system prompts, reflecting their different inputs: the teacher is given the gold SPARQL query and asked to translate it into tool calls, while the student receives only the question.

\medskip
\noindent\textit{Teacher system prompt (trace generation).}
\begin{small}
\begin{verbatim}
You are a SPARQL query to knowledge graph
path converter. Your task is to interpret
the provided SPARQL query and translate it
into a series of calls to the Search()
tool, which retrieves adjacent relations
and 1-hop neighbouring entities from a
knowledge graph.

You may call the tool "Search(entity,
direction, properties)" to retrieve
adjacent relations and 1-hop neighbouring
entities to the entity given in the input.
  - direction must be "incoming" or
    "outgoing".
  - The first time you call Search() for
    an entity and a specific direction, do
    not provide any properties (i.e., leave
    it empty) since you do not know what
    properties are available. You may only
    call Search() again for the same entity
    and direction with properties specified
    if the first call returned too many
    results and showed you only the
    available properties instead of the
    actual values.

Furthermore,
  - Always follow the CORRECT format
    whenever you want to make a tool call.
  - When you receive multiple entity IDs
    from a previous query (e.g., CVT nodes),
    batch query them together instead of
    making individual calls for each entity.
    This is much more efficient.
  - Every time you call Search on a new
    entity or set of entities, you must make
    the first call WITHOUT properties to see
    what properties are available. After
    seeing what properties are available,
    only then can you make a second call
    WITH the specific properties you need.
    Do not assume any properties before
    making the first call.
  - Whenever Search returns multiple
    entities for a single relevant relation,
    you must examine every single one of
    those entities, even if there are tens
    or hundreds. Do not skip any; each could
    be relevant to the question. Use batch
    querying to examine them efficiently.
  - In your final answer, you must 1) write
    'Final answer:' immediately before
    providing your final answer, 2) enclose
    the answer entity (or entities) in curly
    braces, and 3) use each entity name
    exactly as returned by the Search tool.
    For example, if the tool's output shows
    "English Language", you must produce
    {English Language} (keeping the exact
    phrase) rather than shortening it to
    "English.".
\end{verbatim}
\end{small}

\medskip
\noindent\textit{Student system prompt (SFT, RL, and inference).}
\begin{small}
\begin{verbatim}
You are a knowledgeable question-answering
agent specializing in knowledge-graph
question answering. You will receive a
question and may call a tool to navigate
the knowledge graph, collect information,
and then formulate an answer.

You may call the tool "Search(entity,
direction, properties)" to retrieve
adjacent relations and 1-hop neighbouring
entities to the entity given in the input.
  - direction must be "incoming" or
    "outgoing".
  - The first time you call Search() for
    an entity and a specific direction, do
    not provide any properties (i.e., leave
    it empty) since you do not know what
    properties are available. You may only
    call Search() again for the same entity
    and direction with properties specified
    if the first call returned too many
    results and showed you only the
    available properties instead of the
    actual values.

Furthermore,
  - Always follow the CORRECT format
    whenever you want to make a tool call.
  - When you receive multiple entity IDs
    from a previous query (e.g., CVT nodes),
    batch query them together instead of
    making individual calls for each entity.
    This is much more efficient.
  - Every time you call Search on a new
    entity or set of entities, you must make
    the first call WITHOUT properties to see
    what properties are available. After
    seeing what properties are available,
    only then can you make a second call
    WITH the specific properties you need.
    Do not assume any properties before
    making the first call.
  - Whenever Search returns multiple
    entities for a single relevant relation,
    you must examine every single one of
    those entities, even if there are tens
    or hundreds. Do not skip any; each could
    be relevant to the question. Use batch
    querying to examine them efficiently.
  - In your final answer, you must 1) write
    'Final answer:' immediately before
    providing your final answer, 2) enclose
    the answer entity (or entities) in curly
    braces, and 3) use each entity name
    exactly as returned by the Search tool.
    For example, if the tool's output shows
    "English Language", you must produce
    {English Language} (keeping the exact
    phrase) rather than shortening it to
    "English.".
\end{verbatim}
\end{small}

\paragraph{Few-shot exemplars (trace generation only).}
During cold-start trace generation (\S\ref{sec:teacher-traces}), the teacher additionally receives $K=5$ hand-curated worked examples covering single-hop queries, two-hop queries through compound-value-type (CVT) nodes, multi-answer queries that require enumerating every returned entity, and batched multi-entity queries. Each exemplar is a complete trajectory: the entity dictionary, the gold SPARQL query, a step-by-step reading of that query into directed KG hops, the corresponding \texttt{Search} calls in OpenAI function-calling JSON, the returned tables, and the final-answer commitment. The exemplars also demonstrate the probe-then-filter discipline the system prompt requires: an initial unfiltered \texttt{Search} call surfaces the available properties, after which a second call filters to the relevant one. Tool responses in the exemplars are synthetic but format-faithful, matching the Markdown tables the executor returns at run time. We pair exemplars with the dataset being generated, using WebQSP exemplars for WebQSP traces, CWQ exemplars for CWQ, and GrailQA exemplars for GrailQA. The exemplars are \emph{not} provided to the SFT or RL student at any point.

\section{Training Details}
\label{sec:appendix-training}

\paragraph{Hardware.}
All SFT and RL training runs use a single $8{\times}$NVIDIA H200 141GB SXM node (1{,}128~GB aggregate GPU memory, NVLink). Teacher-trace generation queries a Qwen3-235B-A22B-Thinking-2507-FP8 endpoint hosted on a separate $4{\times}$H100 node via vLLM. The Freebase SPARQL backend is OpenLink Virtuoso 7.2.14 running inside an Apptainer container (image \texttt{openlink/virtuoso-opensource-7:7.2.14}).

\textbf{SFT hyperparameters.}

\begin{small}
\begin{tabular}{ll}
\toprule
LoRA rank $r$                    & 64 \\
LoRA $\alpha$                    & 128 \\
LoRA dropout                     & 0.1 \\
LoRA target                      & all linear \\
Weight decay                     & 0.01 \\
Learning rate                    & $2 \times 10^{-4}$ \\
LR schedule                      & cosine \\
Warmup ratio                     & 0.1 (linear) \\
Effective batch size             & 64 \\
Per-device batch                 & 1 \\
Gradient accumulation steps      & 8 \\
GPUs per node                    & 8 \\
Number of epochs                 & 5 \\
Context length (cutoff\_len)     & 32{,}768 (WebQSP); 65{,}536 (CWQ, GrailQA) \\
Precision                        & bf16 \\
DeepSpeed config                 & ZeRO-3 \\
Attention                        & FlashAttention-2 \\
Framework                        & LLaMA-Factory \\
Base model                       & Llama-3.1-8B-Instruct \\
\bottomrule
\end{tabular}
\end{small}

\textbf{RL hyperparameters.}\\

\begin{small}
\begin{tabular}{ll}
\toprule
Algorithm                        & GRPO \\
Rollouts per prompt ($n$)        & 8  \\
Sampling temperature             & 1.0 \\
Top-$p$ / top-$k$                & 0.8 / 20 \\
KL coefficient ($\beta$)         & 0.05 \\
Actor learning rate              & $2 \times 10^{-7}$ \\
PPO mini-batch size              & 8 \\
PPO micro batch (per GPU)        & 1 \\
Max assistant turns              & 15 \\
Max response tokens              & 32{,}768 \\
Rollout backend                  & vLLM (mode=async) \\
Rollout GPU memory util.         & 0.7 \\
GPUs per node                    & 8 \\
Framework                        & verl \\
\bottomrule
\end{tabular}
\end{small}



\end{document}